\documentclass[runningheads]{llncs}
\usepackage[T1]{fontenc}
\usepackage[latin9]{inputenc}
\usepackage{array}
\usepackage{verbatim}
\usepackage{multirow}
\usepackage{amsmath}
\usepackage{graphicx}

\makeatletter

\providecommand{\tabularnewline}{\\}

\makeatother

\begin{document}
\title{Using constraint structure and an improved object detection network
to detect the $12^{th}$ Vertebra from CT images with a limited field
of view for image-guided radiotherapy}
\titlerunning{A constraint feature aided $3D$ detection network}
\author{Yunhe Xie\inst{1} \and Kongbin Kang\inst{2}\and Gregory Sharp\inst{1} \and
David P. Gierga\inst{1} \and Theodore S. Hong\inst{1} \and Thomas
Bortfeld\inst{1}}
\authorrunning{Y. Xie \emph{et al.}}
\institute{Massachusetts General Hospital, Department of Radiation Oncology,
Boston, 02114 USA \\ \email{yxie5@mgh.harvard.edu} \and CVCS, Providence,
RI, 02906, USA\\ \email{kongbin.kang@gmail.com}}
\maketitle
\begin{abstract}
Image guidance has been widely used in radiation therapy. Correctly
identifying the bounding box of the anatomical landmarks from limited
field of views is the key to success. In image-guided radiation therapy
(IGRT), the detection of those landmarks like the 12th vertebra (T12)
still requires tedious manual inspections and annotations; and superior-inferior
misalignment to the wrong vertebral body is still relatively common.
It is necessary to develop an automated approach to detect those landmarks
from images. The challenges of training a model to identify T12 vertebrae
automatically mainly are high shape similarity between T12 and neighboring
vertebrae, limited annotated data, and class imbalance. This study
proposed a novel 3D detection network, requiring only a small amount
of training data. Our approach has the following innovations, including
1) the introduction of an auxiliary network to build constraint feature
map for improving the model's generalization, especially when the
constraint structure is easier to be detected than the main one; 2)
an improved detection head and target functions for accurate bounding
box detection; and 3) an improved loss functions to address the high
class imbalance. Our proposed network was trained, validated and tested
on anotated CT images from 55 patients and demonstrated accurate distinguish
T12 vertebra from its neighboring vertebrae of high shape similarity.
Our proposed algorithm yielded the bounding box center and size errors
of $3.98\pm2.04mm$ and $16.83\pm8.34mm$, respectively. Our approach
significantly outperformed state-of-the-arts Retina-Net3D in average
precision (AP) at IoU thresholds of $0.5$ and $0.35$, with AP increasing
from $0$ to $95.4$ and $0$ to $64.7$, respectively. In summary,
our approach has a great potential to be integrated into the clinical
workflow to improve the safety of IGRT. We implemented our approach
in Python and Keras and will make it available under the open-source
MIT License after acceptance.


\end{abstract}

\section{Introduction }

Volumetric data (e.g. computed tomography (CT), cone-beam computed
tomography (CBCT), $4D$ CT, positron emission tomography, magnetic
resonance imaging, etc.) are commonly used in radiation oncology.
Image-guided radiation therapy (IGRT) utilizes image information to
ensure radiation dose to be delivered precisely and effectively, while
reducing side effects. Correctly identifying anatomical landmarks,
like T12 vertebra, is the key to success. In IGRT, the detection of
such landmarks still requires manual inspections and annotations in
a slice-by-slice manner, which can be tedious. More over, superior-inferior
misalignment to the wrong landmark, particularly in the applications
like CBCT which suffers from high level of noises and a limited field
of view (FOV), is still relatively common. It is necessary to develop
an automated approach to detect those landmarks from images.

Although vertebra structures are highly contracted in CT images, detecting
T12 vertebra from IGRT images remains challenge due to the shape similarity
among the neighboring vertebrae. Automated vertebra detection has
been an active research topic for decades. Previous approaches \cite{HERRING200174,Vrtovec2005,MASTMEYER2006560,Vrtovec2007,Schmidt2007Spine,Klinder2009,Guo2016}utilized
the hand-crafted features to detect vertebrae, which is can be time-consuming,
labor-intensive, and highly dependent on the expert's knowledge and
experience. Moreover, hand-crafted features may not be able to capture
the complex patterns and variations in the data, especially in high-dimensional
and noisy data. Because of the superior performance of Convolutional
Neural Networks (CNN) and their variant, fully convolutional neural
networks (FCN) to learn image features of the structures directly
from data\cite{Ronneberger15UNet,Li2015Cancer,Dou2016Microbleeds,Huang2017Nodule,Haskins2019DeepLI},
it is natural to develop CNN or FCN based approaches. Recently, Cheng
\emph{et al.} \cite{Cheng2021} presented a FCN based method to detect
vertebrae of the full spine. However, their approach focused on full
spine, using a FCN based approach to detect vertebra from limited
view remains an important research topic that needs to be addressed.%
{} 

Although state-of-the-arts detection networks like Retina-Net \cite{Lin2017RetinaNet}
have achieved impressive results in computer vision, they often struggle
to distinguish the T12 vertebra from the neighboring vertebrae of
high similarity. To address this challenge, we propose a novel approach
that introduces an auxiliary network to generate deep-learning features
of thoracic rib 12 (T12 rib) to assist the detection network to distinguish
the T12 from its neighbors. T12 rib is the last pair of ribs in the
human ribcage and is relatively easy to detect from CT scans. It articulates
with T12 vertebra at the back of spine and forms a constraint to T12
vertebra. In addition to the constraint feature network, our approach
differs from state-of-the-art methods by attaching the detection head
to both encoder and decoder paths of the U-Net to improve the detection
accuracy, rather than relying merely on the decoder path of the U-Net. 

In summary, our approach has several innovations. Firstly, we introduce
an auxiliary network to generate deep-learning features assisting
the detection network in improving the model's generalization, especially
when the constraint structure is easier than the main structure to
detect. Secondly, we introduce an improved detection head and target
functions for accurate bounding box detection. Finally, we propose
an improved loss functions to address the high class imbalance. To
evaluate our proposed approach, we trained the network to detect T12
vertebra from CT images. This particular vertebra, which is the last
thoracic vertebra in the spine of the human body, was chosen due to
its technical difficulty in being distinguished from its neighboring
vertebrae in the limited FOV. Additionally, the T12 vertebra is widely
used as a landmark in IGRT and various other radiation oncology clinical
applications. By focusing on the detection of T12, we aimed to demonstrate
the efficacy of our approach in a challenging context that has practical
implications in clinical settings.

\section{Network Architecture}

As shown in Fig.~\ref{fig:Network-Architecture}, our proposed architecture
consists of three components: a 3D main feature extraction network
(MFEN), a 3D constraint feature extraction network (CFEN), and a 3D
region detection network (RDN). RDN is then attached to the top of
both MFEN and CFEN to output the bounding box field.

\subsection{$3D$ FENs for structure to be detected and CFEN for constraint structure}

Our MFEN and CFEN are designed to leverage the effectiveness of 3D
U-Net in handling limited annotated training datasets, which has been
shown in various medical research studies. To achieve this, we use
a standard 3D U-Net with a $4$-layer up-path and down-path as its
backbones. MFEN uses the full down-path and first two layers in up-path,
and CFEN uses only the down-path of U-Net, as shown in the dash box
of Fig.~\ref{fig:Network-Architecture}a. The input to both networks
of MFEN and CFEN is a volumetric image of dimension $N_{x}\times N_{y}\times N_{z}$.
The outputs from CFEN and MFEN down-path feed to a down-path fusion
module. The fusion module first resamples the tensors to a final spatial
resolution $M_{x}\times M_{y}\times M_{z}$ and then concatenate channels
into a final tensor of dimension $M_{x}\times M_{y}\times M_{z}\times d$.
The outputs from MFEN up-path also feed into a fusion module to match
the spatial resolution and concatenate the features into a tensor
of $M_{x}\times M_{y}\times M_{z}\times u$. These two feature fusion
modules combines the multiple layer outputs of the different spatial
resolution, allowing the model to learn the optimal way to select
the correct layer from training data, which is particularly useful
to the structures of anisotropic dimensions. In our experiments, we
used values of $N=416\times216\times128$, $M=52\times36\times32$
for training, with $d$ and $u$ values of 112 and 96, respectively.
The intuition behind this architecture is that CFEN learns the features
of constraint structure and aids MFEN to learn the major features
to achieve a good generalization. 

\subsection{3D Region Detection Network }

RDN attaches to both down-path feature fusion and up-path feature
modules. Its output is a tensor of size $M_{x}\times M_{y}\times M_{z}\times8l$,
where $l$ represents the number of structures to be detected and
8 represents the number of the bounding box parameters (two for probability,
three for locations and three for box sizes). The output tensor represents
a field of box parameters, including the inside-box probability, box
center probability, box location and box size, defined on the coarse-level
spatial grids. In this paper, we use the bounding box aligned to the
image coordinate axis for simplicity. To recover the true box center
from the coarse spatial grids, it is necessary to estimate the the
true center location (or offset). While our network is designed to
detect multiple anatomical structures simultaneously, we only demonstrated
one structure in our experiments.
\begin{figure}[b]
\begin{centering}
\begin{tabular}{cc}
\includegraphics[height=1.2in]{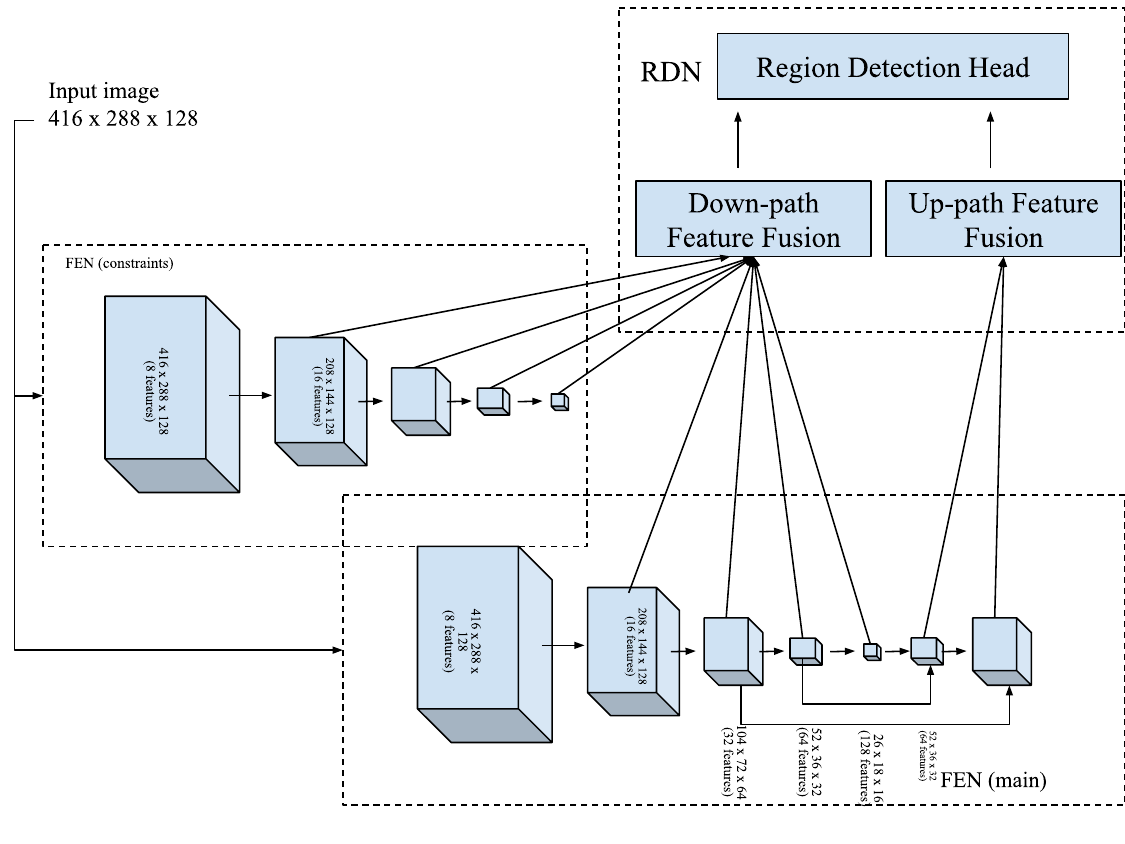} & \includegraphics[height=1.2in]{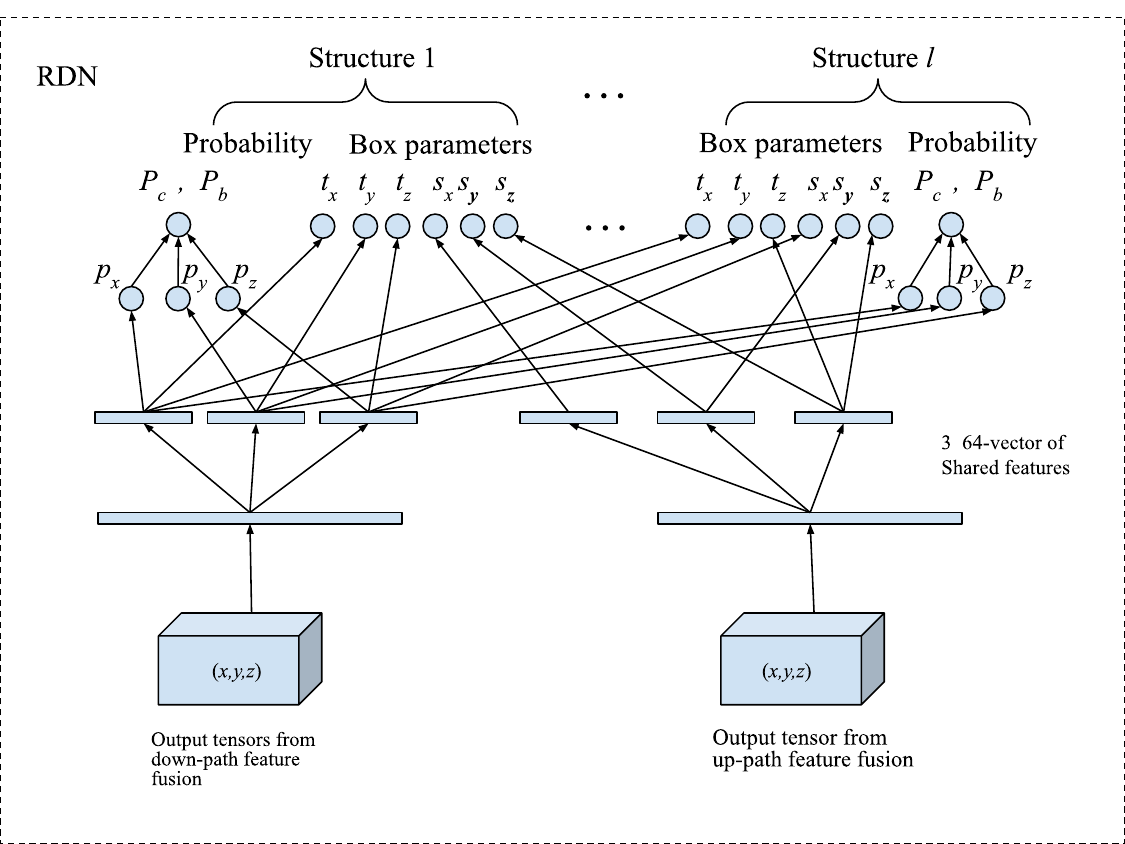}\tabularnewline
a & b\tabularnewline
\end{tabular}
\par\end{centering}
\caption{\label{fig:Network-Architecture}The schematic view of our networks }
\end{figure}

The architecture of the RDN is as shown in Fig.~\ref{fig:Network-Architecture}b.
The input tensors from both fusion module are firstly convolved with
three small $3\times3\times3$ kernels to generate three low dimensional
feature maps, each consisting of 64-vector at a grid point $(x,y,z)$,
represented by narrow bars in Fig.~\ref{fig:Network-Architecture}b).
Next, the resulting feature vectors are convolved by $8l$ $1\times1\times1$-kernels
to generate all the $x$-, $y$- and $z$-components of the inside-box
probability $p_{b}$, box center probability $p_{c}$, location $t$
and box size $s$. Finally, the joint probability, $p$, is a multiplication
from its three axis-components. This architecture was implemented
naturally using 3D convolution and multiplication layers. To mitigate
the effects of image scanning intensity variations, a sigmoid function
is used as the activation function in the convolution layer with $3\times3\times3$-kernels.
Our RDN differs from its counterpart in \cite{Ren2017FasterRCNN}
in that it decomposes the 3D box parameter regressions to multiple
independent regressions on each axis. This approach reduces the number
of weight parameters required in the network, thus helping to mitigate
overfitting with limited training data.

\section{Target Labeling}

\subsection{Augmentation}

Besides the network architecture, data augmentation is another important
step to train deep networks with limited training data sets. A properly
designed augmentation teaches the network to focus on robust features
for a good generalization. In our study, each input image and its
corresponding annotated masks were augmented twenty-five times using
a $2D$ elastic deformation algorithm based on \cite{Ronneberger15UNet}.
In each augmentation, a $3\times3$ grid of random displacements were
drawn from a Gaussian distribution ($\mu=(0,0)$ pixels and $\sigma=(10,10)$
pixels). Then the displacements were interpolated to the pixel level.
All the $2D$ slices of the volumetric image were deformed using the
same displacements and a spline interpolation. 

\subsection{RDN and FEN Output Target}

CFEN was trained using standard U-Net setup and only the down path
was kept after training. We also devised a customized ground truth
approach to train RDN and FEN networks. Although standard target generating
methods, such as the intersection over union (IoU), are commonly used
in computer vision \cite{Ross2015FastRCNN,Ren2017FasterRCNN}, we
found that they were not effective in our experiments due to the optimization
plateau issues\cite{Rezatofighi2019GIoU}. Instead, we developed a
hand-crafted probability function that has a global peak value at
the box center to overcome these problems.

To create the ground truth, we started by obtaining annotated images
and use the structure contours drawn by experienced radiation oncologist
or physicists to compute the target box parameters, including the
box center locations and box sizes. We then used these parameters
to calculate an 8-vector target, $(\hat{p}_{c},\hat{p}_{b}\hat{t}_{x},\hat{t}_{y},\hat{t}_{z},\hat{s}_{x},\hat{s}_{y},\hat{s}_{z})$,
for every spatial grid location $(x,y,z)$. The target was designed
as follows: when inside a target box $b$, the inside-box probability
target $\hat{p}_{b}=1$ and 0 otherwise. The center probability $p_{c}$
is sigmoid decay from the box center, $(x_{bc},y_{bc},z_{bc})$, along
the three axes; the center offset targets, $\hat{t}_{b}$, linearly
increases from the box center; and the box size target, $\hat{s}_{b}$,
is a constant vector. When outside the box $b$, $\hat{p}_{b}$, $\hat{t}_{b}$,
and $\hat{s}_{b}$ are set to zero. This approach teaches the network
to focus on the features of the structure and ignore those outside
it. All those targets are scaled by the down sampling rate, $(r_{x},r_{y},r_{z})$,
caused by pooling operators inside FEN to match the coarse spatial
resolution of RDN output. 

\begin{eqnarray*}
\hat{p}_{c}(x,y,z) & = & \sigma\left(4\left[1-\frac{\left|\Delta x\right|}{r_{x}}\right]\right)\sigma\left(4\left[1-\frac{\left|\Delta y\right|}{r_{y}}\right]\right)\sigma\left(4\left[1-\frac{\left|\Delta z\right|}{r_{z}}\right]\right)\\
p_{b}(x,y,z) & = & \begin{cases}
1 & (x,y,z)\in b\\
0 & otherwise
\end{cases}\\
\hat{t}_{b}(x,y,z) & = & \begin{cases}
\left(\frac{\Delta x}{r_{x}},\frac{\Delta y}{r_{y}},\frac{\Delta z}{r_{z}}\right) & (x,y,z)\in b\\
\left(0,0,0\right) & otherwise
\end{cases}\\
\hat{s}_{b}(x,y,z) & = & \begin{cases}
\left(\frac{w_{b}}{r_{x}},\frac{l_{b}}{r_{y}},\frac{h_{b}}{r_{z}}\right) & (x,y,z)\in b\\
\left(0,0,0\right) & otherwise
\end{cases}
\end{eqnarray*}
where $\sigma(\cdot)$ is the sigmoid function. $\Delta x$, $\Delta y$
and $\Delta z$ are $x-x_{bc}$, $y-y_{bc}$ and $z-z_{bc}$, respectively,
and $w_{b}$, $l_{b}$, and $h_{b}$ are the box width, length and
height, respectively. 

\section{RDN Loss Functions}

We assign a total loss to measure the discrepancy between the RDN
output tensor and the target tensor. The total loss, $L_{total}$,
consists of the losses of the probability $p$, the location offset
$t$, and the box size $s$. Here $p$, $t$, and $s$ are the estimated
probability, center offset and box size, and $\hat{p}$, $\hat{t}$
and $\hat{s}$ are their ground-truth counterparts. The loss functions
are defined as
\begin{eqnarray*}
L_{c} & = & \sum_{b}\frac{\sum_{i,j,k}1_{(i,j,k)\in b}\left\Vert t_{b}(x_{i},y_{j},z_{k})-\hat{t}_{b}(x_{i},y_{j},z_{k})\right\Vert }{\Sigma_{i,j,k}1_{(i,j,k)\in b}}\\
L_{s} & = & \sum_{b}\frac{\sum_{i,j,k}1_{(i,j,k)\in b}\left\Vert s_{b}(x_{i},y_{j},z_{k})-\hat{s}_{b}(x_{i},y_{j},z_{k})\right\Vert }{\Sigma_{i,j,k}1_{(i,j,k)\in b}}\\
L_{p} & = & 1-\frac{2\sum_{i,j,k}\sum_{b}p_{b}(x_{i},y_{j},z_{k})\hat{p}_{b}(x_{i},y_{j},z_{k})}{\sum_{i,j,k}\sum_{b}\left(p_{b}^{2}(x_{i},y_{j},z_{k})+\hat{p}_{b}^{2}(x_{i},y_{j},z_{k})\right)}+\sum_{b}\frac{\sum_{i,j,k}1_{(i,j,k)\in b}\left\Vert p_{c}-\hat{p}_{c}\right\Vert }{\Sigma_{i,j,k}1_{(i,j,k)\in b}}
\end{eqnarray*}
where $i$, $j$ and $k$ are spatial grid index of the RDN output
tensor and $\left\Vert .\right\Vert $ is the $l_{1}-norm$. The total
loss is defined as $L_{total}=L_{p}+L_{c}+L_{s}$. The summation over
all the spatial grid index $(i,j,k)$ is needed in order to measure
the overall discrepancy of the tensor.

With the above definitions, RDN and FEN networks can be trained jointly
end-to-end. The trained network then computes the output tensor when
a tomography image is given. The detection was performed by finding
the maximum probability inside the output tensor and extracting its
associated box parameters. 

\section{Implementation Details }

We implemented our method using Python (ver 3.7.10) with Keras (ver
2.4.3) and TensorFlow (ver 2.4.1) as back-end. We cropped all training
images to 416x288x128 to fit into 24GB GPU memory, and used $(2,2,2)$
pooling size for all pooling layers. Our model was trained end-to-end
using Adam optimizer with a learning rate of $10^{-4}$. It took about
16 hours to train the model on a single NVidia Titan RTX GPU card
with CUDA driver ver 11.4. During the prediction phase, our trained
model generates proposal regions of the top five scores and no structure
is detected if the maximum probability is smaller than 0.5. The training
was run for 150 epochs and the model with the minimum fitting loss
was kept.

\section{Experiment Results}

This study analyzed 55 CT scans from patients who received proton
or photon treatments for various sites including esophagus, lung,
liver, and pancreas at  Masschusset General Hospital from 2012 to
2020. The patients were aged between 34 and 87 years, with 30 males
and 25 females. The CT images were acquired using a GE force CT scanner
at 140 kVp with a field of view ranging from 41 cm to 70 cm. The pixel
sizes ranged from 0.98mm to 1.37mm, and the slice thickness was 2.5mm.
Patients were immobilized on a flat couch in the same posture as for
the radiation treatments during the scanning, and T12 vertebra and
ribs were contoured by experienced radiation oncologists and board-certified
medical physicists. This study was approved by the research IRB.

Those datasets were randomly plit into training, validation and test
datasets consisting of $36$, $5$ and $14$ patients, respectively.
To ensure performance comparability, we implement a 3D RetinaNet with
the same backbone as our proposed algorithm as the baseline. For the
purpose of IGRT safety check, we measured the distance error and the
size error between the ground truth box center and the model predicted
center. Using our trained model, the detection took approximately
3.30 seconds in a CT dataset of $416\times288\times128$ voxels. 

\begin{table}[b]
\begin{centering}
\begin{tabular}{|c|c|c|c|c|}
\hline 
\multicolumn{2}{|c|}{Ablation Studies} & Our approach & RetinaNet (3D) & RetinaNet (3D) + ConstNet\tabularnewline
\hline 
\multirow{2}{*}{AP} & $T=0.35$ & 95.0 & 0 & 41.1\tabularnewline
\cline{2-5} \cline{3-5} \cline{4-5} \cline{5-5} 
 & $T=0.5$ & 64.6 & 0 & 13.7\tabularnewline
\hline 
\end{tabular}
\par\end{centering}
\caption{\label{tab:ap}Ablation studies of the the average precision (AP)
vs introduction of constraint structure (T12 ribs) and improved RDN,
where $T$ represents the IoU threshold.}
\end{table}
On the test CT images, our proposed approach achieved a mean error
of $0.58$, $0.74$, and $1.00$mm in the left-right, anterior-posterior
and superior-inferior directions, respectively, for the detected center.
The corresponding standard deviations of the center location error
were $2.90$, $2.78$, and $3.36$mm, respectively. For the box sizes,
our approach yielded a mean error of $0.64$, $-2.38$, and $-1.40$mm
with std. of $12.52$, $17.76$, and $13.23$mm on the direction left-right,
anterior-posterior and superior-inferior directions, respectively.
We summarized these results in Table~\ref{tab:detection-error}.
Specifically, we accurately identified the center of T12s with a mean
overall detection error of $3.98\pm2.04$mm, and the superior-inferior
mean detection error is $1.00$mm. Given a typical slice thickness
is $2.5$mm, our approach's accuracy is sufficient enough to be integrated
into the IGRT workflow.

\begin{figure}
\begin{centering}
\begin{tabular}{cc}
\includegraphics[width=1.5in]{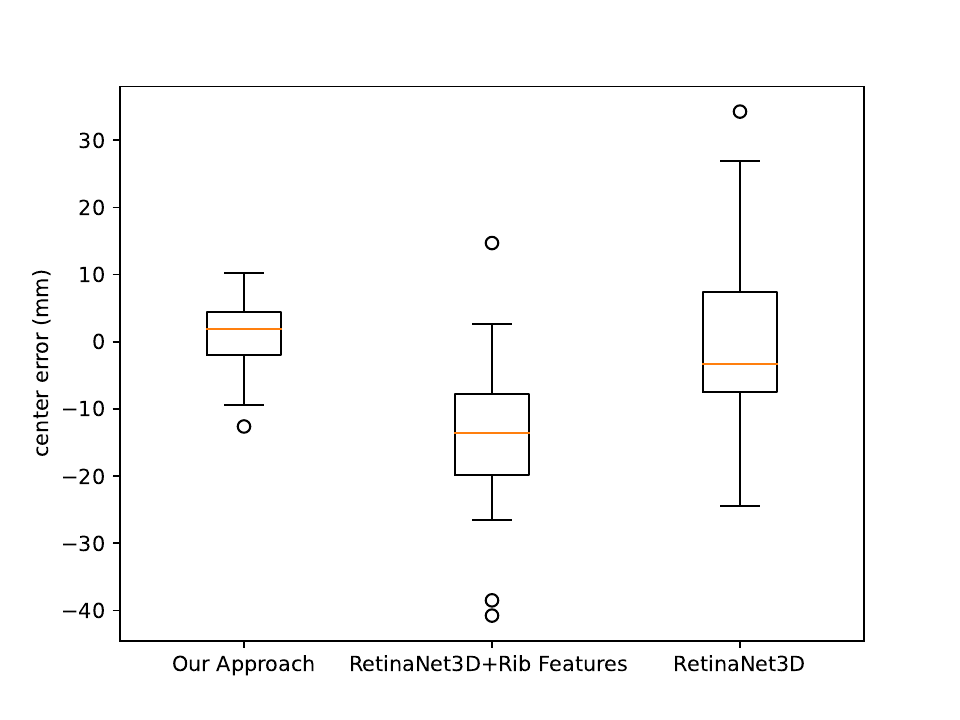} & \includegraphics[width=1.5in]{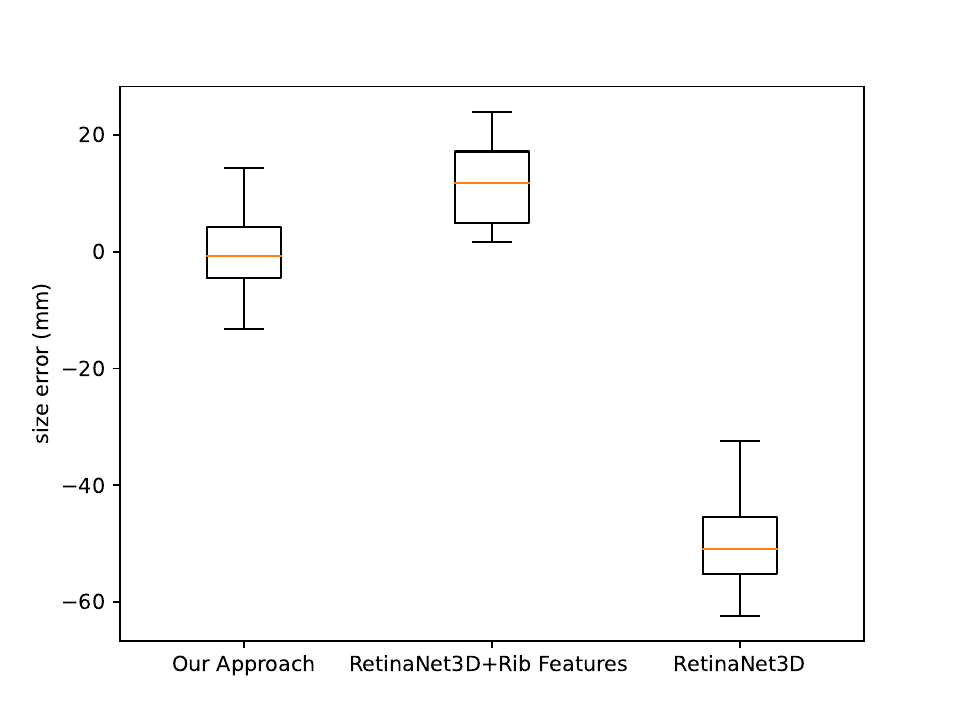}\tabularnewline
\end{tabular}
\par\end{centering}
\caption{\label{fig:comp-rpn}Box plots of the bounding box location and size
errors of the best top 5 detection. left) the box plot of center location
errors. right) the box plot of the box size errors.}
\end{figure}

We conducted ablation studies to compare our approach to the baseline
RetinaNet3D and investigated the detection performance changes with
each proposed improvement. In the study, the matching IoU threshold
for positive proposals was selected as 0.35 and 0.5, which is the
same as the popular choice of 2D \cite{Ren2017FasterRCNN} or an extrapolating
the 3D IoU threshold\cite{Ren2017FasterRCNN} to $3D$ by $\left(\sqrt{0.5}\right)^{3}=0.35$,
respectively. In practice, we found that the detected bounding box
with threshold of $0.35$ already closely matched the ground-truth
. As shown in Table~\ref{tab:ap}, our approach significantly outperformed
the baseline RetinaNet3D in T12 detection AP. Introducing the constraint
feature map improved performance but was still insufficient. However,
incorporating rib constraints and improved RDN significantly enhanced
performance. In the experiments, we found that the RetinaNet3D had
poor T12 detection performance due to detecting the wrong center or
the detection size way off. With the introduction of constraint features
and an improved RDN, our approach consistently and accurately identified
the bounding box, as demonstrated in Fig.~\ref{fig:comp-rpn}. %
\begin{table}[b]
\begin{centering}
\begin{tabular}{|c|c|c|c|c|c|}
\hline 
\multicolumn{2}{|c|}{} & Left-Right  & Anterior-Posterior  & Superior-Inferior & Overall\tabularnewline
\hline 
\multirow{2}{*}{location error $(mm)$} & $\mu$ & 0.58 & 0.74 & 1.0 & 3.98\tabularnewline
\cline{2-6} \cline{3-6} \cline{4-6} \cline{5-6} \cline{6-6} 
 & $\sigma$  & 2.90 & 2.78 & 3.36 & 2.04\tabularnewline
\hline 
\multirow{2}{*}{size error $(mm)$} & $\mu$  & 0.64 & -2.38 & -1.40 & 16.83\tabularnewline
\cline{2-6} \cline{3-6} \cline{4-6} \cline{5-6} \cline{6-6} 
 & $\sigma$ & 12.52 & 17.76 & 13.23 & 8.34\tabularnewline
\hline 
\end{tabular}
\par\end{centering}
\caption{\label{tab:detection-error}Mean and standard deviation of the detection
errors. }
\end{table}

\section{Conclusion}

Our study has presented a novel 3D detection network that incorporates
a constraint feature network, an improved detection head and target
functions, and an improved loss function to accurately detect bounding
boxes of anatomical structures of high similarities in 3D volumetric
data. Our approach outperforms existing methods, with significant
improvement in average precision for different IoU thresholds. With
the ability to train on a small amount of data, our approach has a
great potential to be integrated into the clinical workflow for more
accurate, efficient and safe image guided therapy. Future research
could explore the generalization of the proposed approach to other
anatomical structures and imaging modalities, as well as evaluate
its performance in a larger and more diverse dataset. Furthermore,
we plan to investigate 3D data augmentation techniques to simulate
patients' movements, bending, stretching, and compression etc.

\section{Acknowledgment}

We would also like to thank Dr. Yuling Guo and Dr. Yi Dong for the
technical discussions. 

\bibliographystyle{splncs04}
\bibliography{det20}

\begin{thebibliography}{10}
\providecommand{\url}[1]{\texttt{#1}}
\providecommand{\urlprefix}{URL }
\providecommand{\doi}[1]{https://doi.org/#1}

\bibitem{Cheng2021}
Cheng, P., Yang, Y., Yu, H., He, Y.: Automatic vertebrae localization and
  segmentation in ct with a two-stage dense-u-net. Scientific Reports
  \textbf{11} (12 2021). \doi{10.1038/s41598-021-01296-1}

\bibitem{Dou2016Microbleeds}
{Dou}, Q., {Chen}, H., {Yu}, L., {Zhao}, L., {Qin}, J., {Wang}, D., {Mok},
  V.C., {Shi}, L., {Heng}, P.: Automatic detection of cerebral microbleeds from
  mr images via 3d convolutional neural networks. IEEE Transactions on Medical
  Imaging  \textbf{35}(5),  1182--1195 (May 2016).
  \doi{10.1109/TMI.2016.2528129}

\bibitem{Ross2015FastRCNN}
Girshick, R.: Fast {R-CNN}. In: Proceedings of the 2015 IEEE International
  Conference on Computer Vision (ICCV). pp. 1440--1448. ICCV15, IEEE Computer
  Society, USA (2015). \doi{10.1109/ICCV.2015.169},
  \url{https://doi.org/10.1109/ICCV.2015.169}

\bibitem{Guo2016}
Guo, W., Chen, Q., Zhou, H., Zhang, G., Cong, L., Li, Q.: {Computerized scheme
  for vertebra detection in CT scout image}. In: Tourassi, G.D., III, S.G.A.
  (eds.) Medical Imaging 2016: Computer-Aided Diagnosis. vol.~9785, p. 97853Q.
  International Society for Optics and Photonics, SPIE (2016).
  \doi{10.1117/12.2216744}, \url{https://doi.org/10.1117/12.2216744}

\bibitem{Haskins2019DeepLI}
Haskins, G., Kruger, U., Yan, P.: Deep learning in medical image registration:
  A survey. Machine Vision and Applications  \textbf{31}(8) (2020)

\bibitem{HERRING200174}
Herring, J.L., Dawant, B.M.: Automatic lumbar vertebral identification using
  surface-based registration. Journal of Biomedical Informatics
  \textbf{34}(2),  74--84 (2001). \doi{https://doi.org/10.1006/jbin.2001.1003},
  \url{https://www.sciencedirect.com/science/article/pii/S1532046401910032}

\bibitem{Huang2017Nodule}
{Huang}, X., {Shan}, J., {Vaidya}, V.: Lung nodule detection in {CT} using {3D}
  convolutional neural networks. In: 2017 IEEE 14th International Symposium on
  Biomedical Imaging (ISBI 2017). pp. 379--383 (April 2017).
  \doi{10.1109/ISBI.2017.7950542}

\bibitem{Klinder2009}
Klinder, T., Ostermann, J., Ehm, M., Franz, A., Kneser, R., Lorenz, C.:
  Automated model-based vertebra detection, identification, and segmentation in
  ct images. Medical Image Analysis  \textbf{13},  471--482 (2009).
  \doi{10.1016/j.media.2009.02.004}

\bibitem{Li2015Cancer}
Lee, H., Chen, Y.P.P.: Image based computer aided diagnosis system for cancer
  detection. Expert Syst. Appl.  \textbf{42}(12),  5356--5365 (July 2015)

\bibitem{Lin2017RetinaNet}
Lin, T., Goyal, P., Girshick, R.B., He, K., Doll{\'{a}}r, P.: Focal loss for
  dense object detection. CoRR  \textbf{abs/1708.02002} (2017),
  \url{http://arxiv.org/abs/1708.02002}

\bibitem{MASTMEYER2006560}
Mastmeyer, A., Engelke, K., Fuchs, C., Kalender, W.A.: A hierarchical 3d
  segmentation method and the definition of vertebral body coordinate systems
  for qct of the lumbar spine. Medical Image Analysis  \textbf{10}(4),
  560--577 (2006). \doi{https://doi.org/10.1016/j.media.2006.05.005},
  \url{https://www.sciencedirect.com/science/article/pii/S1361841506000363},
  special Issue on Functional Imaging and Modelling of the Heart (FIMH 2005)

\bibitem{Ren2017FasterRCNN}
{Ren}, S., {He}, K., {Girshick}, R., {Sun}, J.: Faster {R-CNN}: Towards
  real-time object detection with region proposal networks. IEEE Transactions
  on Pattern Analysis and Machine Intelligence  \textbf{39}(6),  1137--1149
  (June 2017). \doi{10.1109/TPAMI.2016.2577031}

\bibitem{Rezatofighi2019GIoU}
Rezatofighi, S.H., Tsoi, N., Gwak, J., Sadeghian, A., Reid, I.D., Savarese, S.:
  Generalized intersection over union: {A} metric and {A} loss for bounding box
  regression. CoRR  \textbf{abs/1902.09630} (2019),
  \url{http://arxiv.org/abs/1902.09630}

\bibitem{Ronneberger15UNet}
Ronneberger, O., Fischer, P., Brox, T.: U-net: Convolutional networks for
  biomedical image segmentation. In: Medical Image Computing and
  Computer-Assisted Intervention (MICCAI). LNCS, vol.~9351, pp. 234--241.
  Springer (2015),
  \url{http://lmb.informatik.uni-freiburg.de/Publications/2015/RFB15a},
  (available on arXiv:1505.04597 [cs.CV])

\bibitem{Schmidt2007Spine}
Schmidt, S., Kappes, J., Bergtholdt, M., Pekar, V., Dries, S., Bystrov, D.,
  Schn{\"o}rr, C.: Spine detection and labeling using a parts-based graphical
  model. Inf. Process. Med. Imaging  \textbf{20},  122--133 (2007)

\bibitem{Vrtovec2005}
Vrtovec, T., Likar, B., Pernuš, F.: Automated curved planar reformation of 3d
  spine images. Physics in Medicine \& Biology  \textbf{50}(19), ~4527 (sep
  2005). \doi{10.1088/0031-9155/50/19/007},
  \url{https://dx.doi.org/10.1088/0031-9155/50/19/007}

\bibitem{Vrtovec2007}
Vrtovec, T., Ourselin, S., Gomes, L., Likar, B., Pernuš, F.: Automated
  generation of curved planar reformations from mr images of the spine. Physics
  in Medicine \& Biology  \textbf{52}(10), ~2865 (apr 2007).
  \doi{10.1088/0031-9155/52/10/015},
  \url{https://dx.doi.org/10.1088/0031-9155/52/10/015}

\end{thebibliography}

\end{document}